\pdfoutput=1

\documentclass[11pt]{article}

\usepackage[preprint]{acl}

\usepackage{times}
\usepackage{latexsym}

\usepackage[T1]{fontenc}

\usepackage[utf8]{inputenc}

\usepackage{microtype}

\usepackage{inconsolata}

\usepackage{graphicx}

\usepackage{enumitem}
\usepackage{amsmath}
\usepackage{booktabs}
\usepackage{subfigure}
\usepackage{multirow}
\usepackage{xcolor}         
\usepackage{mdframed}

\definecolor{customblue}{HTML}{ADD8E6} 

\usepackage{colortbl}
\newcommand{\equalcontrib}{\textsuperscript{\textdagger}}  
\newcommand{\correspondauthor}{\textsuperscript{*}}

\newcommand{\ul}[1]{\underline{#1}}

\title{\fontsize{13.5}{15}\selectfont{ToolACE-DEV: Self-Improving {Tool} Learning via \ul{D}ecomposition and \underline{EV}olution}}

\author{
 \textbf{Xu Huang\textsuperscript{1}}\equalcontrib,
 \textbf{Weiwen Liu\textsuperscript{2}}\equalcontrib,
 \textbf{Xingshan Zeng\textsuperscript{2}},
 \textbf{Yuefeng Huang\textsuperscript{1}},\\
 \textbf{Xinlong Hao\textsuperscript{2}},
 \textbf{Yuxian Wang\textsuperscript{3}},
 \textbf{Yirong Zeng\textsuperscript{3}},
 \textbf{Chuhan Wu},\\
 \textbf{Yasheng Wang\textsuperscript{2}}\correspondauthor,
 \textbf{Ruiming Tang\textsuperscript{2}},
 \textbf{Defu Lian\textsuperscript{1}}\correspondauthor
\\
\\
 \textsuperscript{1}University of Science and Technology of China,
 \\
 \textsuperscript{2}Huawei Noah's Ark Lab,
 \textsuperscript{3}Huawei Technologies Co., Ltd
 \\
 \small{
     \href{mailto:xuhuangcs@mail.ustc.edu.cn}{xuhuangcs@mail.ustc.edu.cn}
     \href{mailto:wwliu@sjtu.edu.cn}{wwliu@sjtu.edu.cn}
 }
}

\begin{document}
\maketitle
\begin{abstract}
The tool-using capability of large language models (LLMs) enables them to access up-to-date external information and handle complex tasks. Current approaches to enhancing this capability primarily rely on distilling advanced models by data synthesis. However, this method incurs significant costs associated with advanced model usage and often results in data compatibility issues, led by the high discrepancy in the knowledge scope between the advanced model and the target model.
To address these challenges, we propose \textbf{ToolACE-DEV}, a self-improving framework for tool learning. First, we decompose the tool-learning objective into sub-tasks that enhance basic tool-making and tool-using abilities. Then, we introduce a self-evolving paradigm that allows lightweight models to self-improve, reducing reliance on advanced LLMs. Extensive experiments validate the effectiveness of our approach across models of varying scales and architectures.
\end{abstract}

{
\renewcommand{\thefootnote}{\textdagger}
\footnotetext[1]{Equal Contributions.}
\renewcommand{\thefootnote}{*}
\footnotetext[2]{Corresponding authors.}
\renewcommand{\thefootnote}{\fnsymbol{footnote}}
}

\section{Introduction}
Large Language Models (LLMs) have achieved remarkable progress in natural language processing. However, they face significant limitations, including factual inaccuracies and challenges in accessing real-time information or executing actions. Enhancing their ability to use external tools—such as search engines~\cite{schick2023toolformer,nakano2021webgpt}, APIs~\cite{qin2023toolllm}, and mathematical tools~\cite{cobbe2021training,he2023solving}—is a promising solution. Tool integration not only grounds LLMs' outputs in reliable information but also expands their applicability to real-world scenarios requiring complex interactions.

Existing approaches to improve tool-utilization capabilities typically rely on distilling advanced models like GPT-4 or Claude 3.5 through data synthesis~\cite{patil2023gorilla,qin2023toolllm,tang2023toolalpaca,lin2024hammer,liu2024toolacewinningpointsllm}. However, this strategy introduces three major challenges:
1) \textit{Inference Cost.} Utilizing advanced models is prohibitively expensive, particularly when generating large-scale training datasets.
2) \textit{Data Compatibility.} The synthesized data frequently exhibits distributional discrepancies, making it less compatible with the target model being fine-tuned. Specifically, unfamiliar samples—those introducing concepts outside the base model's knowledge scope—often lead to hallucinations~\cite{hartmann2023sok,kang2024unfamiliar}.
Consequently, target models tend to memorize the training data rather than generalize from it~\cite{tirumala2022memorization,setlur2024rl}, ultimately leading to suboptimal tool-utilization performance.
3) \textit{Data Privacy.} In real-world applications, numerous user queries involve privacy constraints, prohibiting the synthesis using external advanced models.
A promising alternative is self-evolution~\cite{tao2024survey}, where a model generates or refines its own training data, enabling iterative improvement without heavy reliance on external resources. Self-evolution has shown success in enhancing reasoning~\cite{gulcehre2023reinforced,singh2023beyond,huang2023selfimprove} and code-generation~\cite{jiang2023selfevolve,chen2024teaching} tasks through techniques like top-k sampling or nucleus sampling, where multiple solutions are generated, and correct ones are used for fine-tuning. 


However, applying self-evolution to iterative improvement in tool-learning scenarios presents unique challenges. Tool-learning tasks typically consist of three components: user queries, candidate tools, and ground-truth tool invocations.
Enhancing models with a diverse range of candidate tools during fine-tuning has been shown to improve their overall proficiency and zero-shot capabilities in tool utilization~\cite{liu2024toolacewinningpointsllm}. 
While lightweight, open-source LLMs demonstrate the ability to invoke tools from predefined candidate sets, they struggle to generate both novel tools and accurate invocations directly from user queries~\cite{huang2024planning}. 
This limitation makes the generation of high-quality, diverse training data a significant challenge.


To address the aforementioned challenges, we first \textbf{decompose} tool learning into several tool-related sub-tasks, enhancing the tool-making and tool-using abilities. Then we propose a \textbf{self-evolution} strategy specifically tailored for tool-learning, enabling the model to self-improve. The overall pipeline is termed as \textbf{ToolACE-DEV}. First, we identified that constructing tool documentation adaption tasks focused on tool definitions for post-trained models can effectively enhance the model's understanding of tools, thereby improving its tool-using and tool-generation capabilities. Subsequently, we decomposed the conventional tool-learning training objective, which typically concentrates solely on tool-using ability, into two tasks: \textit{tool generation} and \textit{tool invocation}. This approach strengthens the target model's ability to generate candidate tools based on a query, while simultaneously improving the accuracy of tool invocation, equipping the model with the foundational capabilities for self-evolution. Finally, after the aforementioned two-stage training, by providing new user queries, the target model iteratively generates candidate tools and corresponding invocations. This iterative process establishes a self-evolutionary mechanism that automatically enhances the model's tool-utilization performance over time. Our contributions can be summarized as follow:
\begin{itemize}[leftmargin=0pt]
    \item We propose ToolACE-DEV, the first self-evolutionary framework designed to enhance LLMs' tool-invocation capabilities, equipping lightweight models with self-evolving abilities.
    \item We propose the tool documentation adaption sub-task and decompose the tool-learning objective into tool generation and invocation tasks, demonstrating the task decomposition significantly improves tool-invocation performance.
    \item Through extensive experiments on LLMs of varying scales, we validate the effectiveness of our approach and provide insights into how self-evolution potential varies with model size.
\end{itemize}


\begin{figure*}[t]
    \centering
    \includegraphics[width=0.9\linewidth]{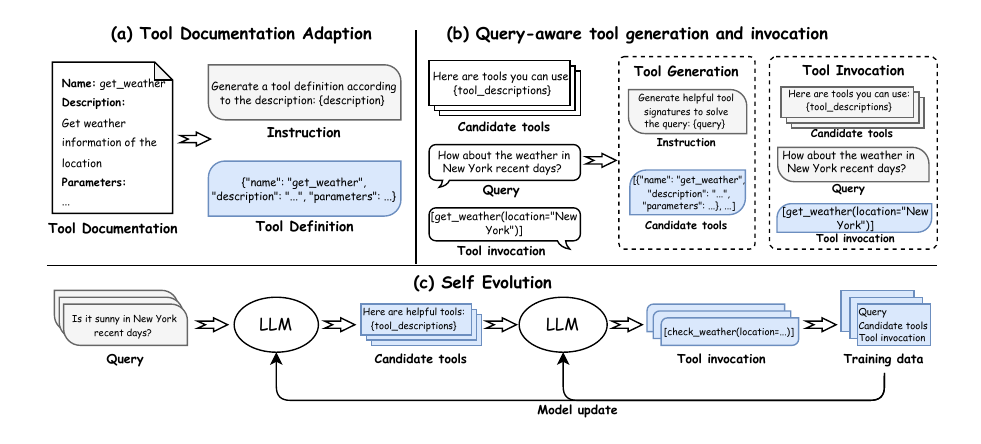}
    \caption{Overall framework of the self-evolving paradigm. (a) Tool documentation adaption, aiming to enhance the understanding of tools; (b) Query-aware tool generation and invocation, equipping the model with self-evolving abilities; (c) Self-evolution, where the model first generates candidate tools and then generates tool invocation, forming the self-training data.}
    \label{fig:framework}
\end{figure*}

\section{Related Work}
\subsection{Tool Learning}

The integration of external tools significantly enhances the capabilities of large language models (LLMs), enabling them to perform more specialized, accurate, and reliable problem-solving tasks~\citep{qin2023toolllm}. Existing methods for equipping LLMs with tool-use capabilities can be broadly categorized into two types: prompt-based approaches and tool-augmented tuning.
Prompt-based methods enable LLMs to use tools by providing in-context examples and tool descriptions, bypassing the need for additional model training~\citep{mialon2023augmented,hsieh2023tool,ruan2023tptu}. A notable example is the ReAct framework~\citep{yaoreact}, which allows LLMs to alternate between reasoning and executing actions to solve complex tasks. While tuning-free methods are lightweight and flexible, their performance is heavily reliant on the LLM’s intrinsic capabilities, which limits their effectiveness for tasks requiring advanced tool utilization.
In contrast, tool-augmented tuning methods directly enhance LLMs’ tool-use capabilities through additional training~\citep{qin2023toolllm,schick2023toolformer,patil2023gorilla,tang2023toolalpaca,liu2024apigen,abdelaziz2024granite,liu2024toolacewinningpointsllm,lin2024hammer}. These methods typically involve fine-tuning LLMs to use external APIs and tools. However, a common limitation is the demand for high-quality data, which highly relies on data synthesis by an advanced model, such as GPT-4 or Claude-3.5. This generation process is not only resource-intensive but also incurs significant costs.

\subsection{Self Evolution of LLMs}
Self-evolution enables models to acquire and update knowledge autonomously, akin to human learning.
For instance, the transition from AlphaGo~\cite{silver2016mastering} to AlphaZero~\cite{silver2017mastering} utilized a self-play mechanism to facilitate model evolution without reliance on labeled data.
In the context of LLM self-evolution, research often focuses on two stages: task acquisition and solution generation. During the task acquisition stage, the target model generates new tasks. For example, Self-Instruct~\cite{wang2022self} enables models to autonomously generate new instructions as tasks, while Ada-Instruct~\cite{cui2023ada} proposes an adaptive approach for task instruction generation. WizardLM~\cite{xu2023wizardlm} introduces the Evol-Instruct method, which evolves instructions through both depth and breadth.
In the solution generation stage, emphasis is placed on producing suitable answers for tasks. The STaR~\cite{zelikman2024star} framework incorporates the model’s reasoning process and uses correct problem-solving steps as training data. REST~\cite{gulcehre2023reinforced} and $\text{REST}^{em}$~\cite{singh2023beyond} employ sampling strategies to generate multiple trajectories and leverage a reward model to guide updates. Other approaches utilize both positive and negative sample pairs for preference-based training, such as DPO~\cite{rafailov2024direct}. Self-Reward~\cite{yuan2024self}, for instance, constructs preference pairs by using the model itself as a reward model after solution generation. SPIN designates model-generated data as negative samples and labeled SFT data as positive samples. GRATH~\cite{chen2024grath} explicitly generates both positive and negative samples simultaneously, while Self-Contrast~\cite{zhang-etal-2024-self-contrast} compares differences between solutions and compiles these differences into a checklist for iterative refinement.
In this work, we implement both task acquisition and solution generation, achieving completely autonomous evolution for LLMs.

\section{Methodology}
\subsection{Task Definitions}
Given the user query $q$ and candidate tools $T=\{t_1, t_2, \cdots, t_N\}$, the goal of the tool invocation task is to select suitable tools and extract information as arguments $A$ with the model parameters $\Theta$:
\begin{equation*}
    A = [\cdots, (t_j, a_j), \cdots] = f(q, T, \Theta)
\end{equation*}
where $t_j$ and $a_j$ represent the $j$-th called tool and corresponding arguments, respectively. $f(\cdot)$ denotes the auto-regressive generation manner of LLMs. The training sample tailored for tool-using is usually formalized as a triplet of a user query, candidate tools and the ground-truth answers: $\langle q, T, A\rangle$.

The overall framework of ToolACE-DEV comprises three stages: tool documentation adaption, query-aware tool generation and innovation, and self-evolution, which is illustrated in Figure~\ref{fig:framework}.

\subsection{Tool Documentation Adaption}
To enhance the LLM's capability on specific domains, continual pre-training on domain data is usually adopted as an effective method~\cite{wu2023bloomberggpt,singhal2025toward}. Drawing inspiration from this, we propose to train LLM on tool documentation for better tool understanding. This process enables the model to acquire a more in-depth understanding of the syntax, functionality, and constraints of specific tools, which can significantly improve its utility in real-world applications. Unlike general-purpose pre-training, this approach equips the model with domain-specific knowledge directly related to tool usage, reducing the gap between training data and deployment scenarios. Unlike other methods that continually pre-train a ``base''-series LLM without any post-training such as instruction tuning, we design a adaption task for ``Instruct''-series models, thereby keeping the instruct-following ability obtained in post-training.

Specifically, given the documentation of a tool: $t_i = \langle name, description, parameters\rangle$, we construct an instruction $x_{t_i}$ to ask the model to complete tool definitions according to tool description, such as ``\textsl{Generate a tool signature according to the description: \{description\}}''. Then the training procedure aligns with an instruction tuning:
\begin{equation}
    \min_\Theta \ell\left(f\left(x_{t_i}, \Theta\right), t_i\right)
\end{equation}
where $\ell(\cdot)$ is the loss function to align the model's prediction with the tool's documentation. The task enables the LLM to be familiar with the format of tool definitions, building the fundamental ability to construct tools for coming queries.

\subsection{Query-Aware Tool Generation and Invocation}\label{sec:stage2}
Unlike question-answering or coding tasks, which typically involve only queries and answers, tool invocation data often requires not only the user's query but also a set of candidate tools. Existing studies generally rely on a finite set of candidate tools sampled from a fixed pool. However, this approach overlooks a critical issue: when models are applied to new scenarios with unseen candidate tools and queries, the accuracy of tool invocation tends to suffer significantly. Therefore, a tool invocation model with self-evolution capabilities should ideally possess the ability to expand its training set of tools autonomously.

To address this challenge, we decompose the tool learning data into two sub-tasks: query-aware tool generation and tool invocation. In existing tool-learning training, the focus is typically on the tool invocation, i.e., given a query and a set of candidate tools, aligning the model's predicted tool invocation with the ground truth $A$:
\begin{equation}\label{eq:tool_invocation}
    \min_\Theta \ell\left(f\left(q, T, \Theta\right), A\right)
\end{equation}

Contrastively, our decomposition introduces an additional task during training: generating query-relevant candidate tools based on the given query. This aims to both enhance the model's understanding of the relationship between queries and tools and equip it with the ability to autonomously generate relevant candidate tools, thereby preparing it for self-evolution. Similar to tool documentation adaption, given a query $q$, we convert it into an instruction format $x_q$, such as ``\textsl{Generate candidate tools related to the query: \{query\}.}'' The training objective is then to generate the corresponding candidate tools $T$:
\begin{equation}\label{eq:tool_generation}
    \min_\Theta \ell\left(f\left(x_q, \Theta \right), T\right)
\end{equation}
By training with this objective, the model becomes capable of generating candidate tools for incoming queries, transcending the limitations of a finite tool set, thereby opening the door to self-evolution.

\subsection{Self-Evolution} \label{sec:self-evol}

After training through the first two stages, the model acquires the foundational capabilities for self-evolution: the ability to generate candidate tools based on a given query, and the ability to invoke tools based on the query and its corresponding candidate tools. When confronted with a new query, the model can autonomously generate new tool invocation training data. The self-evolution process is primarily composed of three steps: candidate tool generation, tool invocation generation, and model updating, illustrated in Figure~\ref{fig:framework}(c).

\textbf{Candidate tool generation.}
Upon a new query $q$ is collected, it is first reformulated into an instruction $x_q$ to guide the model in generating a set of candidate tools $\Tilde{T}$ relevant to the query:
\begin{equation}
    \tilde{T} = f(x_q, \Theta^{(i)})
\end{equation}
To ensure the correctness of the format of generated tools, we adopt a rule checker to filter out those problematic samples, such as missing argument descriptions or JSON-unparsable.

\begin{table*}[thb]
\caption{Accuracy performance comparison on BFCL leaderboard. The top 20 models are listed for comparison. The models are sorted according to the overall score. FC denotes the model is tailored for functional calling.}
\label{tab:overall}
\footnotesize
\setlength{\tabcolsep}{2.5pt}
\renewcommand{\arraystretch}{1.2}
\centering
\begin{tabular}{@{}lccccccccccc@{}}
\toprule
\multirow{3}{*}{\textbf{Model}}& \multicolumn{4}{c}{\textbf{Non-Live}}          & \multicolumn{4}{c}{\textbf{Live}}              & \multicolumn{3}{c}{\textbf{Overall}} \\ \cmidrule(lr){2-5} \cmidrule(lr){6-9} \cmidrule(lr){10-12}
 &
  \multicolumn{1}{l}{\textit{Simple}} &
  \multicolumn{1}{l}{\textit{Multi}} &
  \multicolumn{1}{l}{\textit{Parallel}} &
  \multicolumn{1}{l}{\begin{tabular}[c]{@{}c@{}}\textit{Parallel}\\ \textit{Multi}\end{tabular}} &
  \multicolumn{1}{l}{\textit{Simple}} &
  \multicolumn{1}{l}{\textit{Multi}} &
  \multicolumn{1}{l}{\textit{Parallel}} &
  \multicolumn{1}{l}{\begin{tabular}[c]{@{}c@{}}\textit{Parallel}\\ \textit{Multi}\end{tabular}} &
  \multicolumn{1}{l}{\textit{Non-live}} &
  \multicolumn{1}{l}{\textit{Live}} &
  \multicolumn{1}{l}{\textit{Overall}} \\ \midrule
GPT-4-turbo (Prompt)              & 82.25 & 94.50 & 95.00 & 93.50 & 78.68 & 83.12 & 81.25 & 75.00 & 91.31 & 82.09 & 86.70 \\
xLAM-8x22b-r (FC)                            & 77.00 & 95.50 & 92.50 & 94.00 & 70.93 & 77.72 & 75.00 & 75.00 & 89.75 & 76.33 & 83.04 \\
\rowcolor{customblue} ToolACE-DEV(FC) & 80.17 & 97.50 & 93.50 & 87.50 & 70.16 & 76.37 & 81.25 & 75.00 & 89.67 & 75.20 & 82.44\\
Llama-3-70B-Instruct (Prompt)           & 75.83 & 94.50 & 91.50 & 87.00 & 69.77 & 78.01 & 75.00 & 66.67 & 87.21 & 76.18 & 81.69 \\
mistral-large (FC)                      & 57.50 & 94.00 & 93.00 & 92.00 & 79.07 & 78.88 & 87.50 & 75.00 & 84.12 & 78.95 & 81.54 \\
xLAM-8x7b-r (FC)                             & 77.25 & 95.50 & 92.00 & 89.00 & 68.99 & 76.18 & 50.00 & 75.00 & 88.44 & 74.46 & 81.45 \\
ToolACE-8B (FC) & 80.58 & 95.00 & 91.00 & 90.50 & 62.79 & 74.25 & 81.25 & 75.00 & 89.27 & 72.13 & 80.70\\
GPT-4o-mini (Prompt)              & 79.67 & 89.50 & 89.00 & 88.00 & 72.09 & 73.77 & 81.25 & 70.83 & 86.54 & 73.48 & 80.01 \\
GPT-3.5-Turbo (FC)                      & 74.08 & 93.00 & 87.50 & 83.50 & 65.50 & 74.16 & 56.25 & 54.17 & 84.52 & 71.91 & 78.22 \\
FireFunction-v2 (FC)                         & 78.83 & 92.00 & 91.00 & 81.00 & 69.38 & 70.97 & 56.25 & 54.17 & 85.71 & 70.18 & 77.95 \\
GPT-4-turbo (FC)                  & 60.58 & 91.00 & 90.00 & 89.00 & 67.83 & 74.45 & 75.00 & 62.50 & 82.65 & 72.96 & 77.81 \\
GPT-4o (FC)                       & 73.58 & 92.50 & 91.50 & 84.50 & 67.83 & 69.43 & 75.00 & 66.67 & 85.52 & 69.14 & 77.33 \\
GPT-4o-mini (FC)                  & 67.83 & 90.50 & 89.50 & 83.50 & 67.83 & 69.82 & 81.25 & 70.83 & 82.83 & 69.59 & 76.21 \\
Gorilla-OpenFunctions-v2 (FC)                & 77.67 & 95.00 & 89.00 & 87.50 & 63.95 & 63.93 & 62.50 & 45.83 & 87.29 & 63.59 & 75.44 \\
xLAM-7b-fc-r (FC)                            & 77.33 & 92.50 & 91.50 & 86.00 & 63.57 & 63.36 & 56.25 & 50.00 & 86.83 & 63.08 & 74.95 \\
Open-Mistral-Nemo (FC)                  & 60.92 & 92.00 & 85.50 & 85.50 & 68.22 & 67.98 & 75.00 & 62.50 & 80.98 & 68.01 & 74.50 \\
GPT-4o (Prompt)                   & 64.08 & 86.50 & 88.00 & 85.00 & 67.44 & 67.21 & 56.25 & 58.33 & 80.90 & 66.96 & 73.93 \\
Gemini-1.5-Flash-Preview (FC)           & 65.42 & 94.50 & 71.50 & 77.00 & 62.79 & 72.61 & 56.25 & 54.17 & 77.10 & 70.18 & 73.64 \\
Claude-3.5-Sonnet (FC)              & 75.42 & 93.50 & 62.00 & 50.50 & 72.48 & 70.68 & 68.75 & 75.00 & 70.35 & 71.08 & 70.72 \\
Gemini-1.5-Pro-Preview (FC)             & 50.17 & 89.50 & 83.50 & 79.00 & 60.08 & 66.35 & 75.00 & 54.17 & 75.54 & 65.02 & 70.28 \\
o1-mini (Prompt)                  & 68.92 & 89.00 & 73.50 & 70.50 & 62.79 & 65.09 & 68.75 & 58.33 & 75.48 & 64.57 & 70.02 \\
\bottomrule
\end{tabular}
\end{table*}

\textbf{Tool invocation generation.}
After the generation of candidate tools, the model is then prompted to generate tool calls $\tilde{A}$ to solve the query with generated tools $\tilde{T}$. To improve the correctness of the generated solution, we obtain multiple solutions via the top-k sampling strategy and then majority voting is applied to select the answer as the ground truth. The sampling and voting process, termed as self-consistency decoding~\cite{wang2023selfconsistency}, has been validated as an effective method to improve the performance of LLMs. From the perspective of self-rewarding, it assigns a positive reward to the solution with higher confidence:
\begin{equation}
    \tilde{A} = \operatorname{majority\_vote}(f_{sampling}(q, \tilde{T}, \Theta^{(i)})
\end{equation}
where $\operatorname{majority\_vote}(\cdot)$ denotes select the solutions with the most votes and $f_{sampling}$ denotes the sampling-based decoding strategy, which is implemented as top-k sampling in our experiments. Also, a rule checker is applied to filter out those samples with unreasonable solutions, such as calling hallucinating tools or arguments and filling arguments with wrong types.

\textbf{Model updating.}
For each incoming query, candidate tool generation and tool invocation generation can turn the query $q$ to a complete tool-using triplet $\langle q, \tilde{T}, \tilde{A}\rangle$. Then a new training set can be collected after repeating the first two steps on all queries, where the model can be trained with two types of objectives: query-aware tool generation and invocation, as proposed in Section~\ref{sec:stage2}:
\begin{equation}
    \small
    \Theta^{(i+1)} = \min_\Theta \ell\big(f(q, \tilde{T}, \Theta), \tilde{A}\big) + \ell\big(f\left(x_q, \Theta \right), \tilde{T}\big)
\end{equation}

\section{Experiments}
\subsection{Experimental Settings}
\noindent \textbf{Datasets Construction}.
In the first phase of training, we sampled a subset of data from ToolACE~\cite{liu2024toolacewinningpointsllm}, comprising a total of 26,522 tools, to perform tool documentation adaption on the model. Subsequently, in the second phase, we utilized a dataset containing 20,000 synthesized tool invocation samples generated by GPT-4~\footnote{\url{https://chatgpt.com}} for further fine-tuning. In each subsequent self-evolution round, the model self-generates training data by processing 10,000 incoming queries. We set the max round of self-evolution as 3 and the best results (may not be at the third round) are adopted in Table~\ref{tab:overall}. 
Note that the overall training utilizes 20,000 synthesized query-solution pairs and a maximum of 30,000 queries in total.

\noindent \textbf{Benchmark and Evaluation}.
To evaluate the model's tool invocation capabilities, we selected the Berkeley Function Call Leaderboard (BFCL)~\cite{berkeley-function-calling-leaderboard}, a widely recognized benchmark, as the evaluation framework. BFCL consists of two subsets: Non-live and Live, representing synthetic test cases and real-world scenarios, respectively~\footnote{We focused exclusively on single-turn tool invocation AST data, as these test cases exhibit higher stability and reliability, whereas other cases tend to have significant variability and lower reliability.}. Both non-live and live subsets comprise four types of test examples: simple, multiple, parallel, and parallel multiple.  Simple and multiple examples both involve only one invoked tool, while there are multiple candidate tools in multiple examples. Parallel (or Parallel multiple) examples require invoking multiple tools from one (or multiple) candidate tool(s). The evaluation metrics for each subset are accuracy-based, and for certain categories, the scores are computed as the average of subcategory scores. 
To further validate the efficiency of our method, we evaluate ToolACE-DEV on another two tool-calling benchmarks: API-Bank~\cite{li2023api} and T-Eval~\cite{chen-etal-2024-eval}.
The details of those benchmarks are reported in Appendix~\ref{sec:appd_exp}.

\begin{table}[t]
    \centering
    \caption{Results on other tool learning benchmarks. \textbf{Bold} and \underline{underline} results represent the 1st and the 2nd best results.}
    \begin{tabular}{l|cc}
        \toprule
        \textbf{Model} & \textbf{APIBank} & \textbf{T-Eval} \\
        \midrule
         GPT-4-turbo & \underline{63.39} & \textbf{87.50} \\
         Llama-3.1-8B-Instruct & 54.11 & 76.60 \\ \midrule
         \textbf{ToolACE-DEV} & \textbf{67.82} & \underline{77.03} \\
         \bottomrule
    \end{tabular}
    \label{tab:overall_other}
\end{table}

\noindent \textbf{Implementation Details}.
We employed the LLaMA3.1-8B-Instruct~\cite{llama3modelcard} as the base model for training. Due to resource constraints, the parameter-efficient training technique, LoRA~\cite{hu2022lora}, is conducted on 8 Nvidia V100-32GB GPUs. All model modules are enabled for LoRA fine-tuning, with the LoRA rank set to 16 and alpha set to 32. The training processes utilize a global batch size of 64 and a learning rate of $10^{-4}$ with a commonly used cosine learning rate scheduler, where the warmup ratio is set as 0.1. The prompts in each stage are illustrated in Appendix~\ref{sec:appd_prompt}. More details are provided in Appendix~\ref{sec:appd_exp}.

\subsection{Main Results}

To demonstrate the superiority of the model performance under the training framework we propose, we compare the tool invocation accuracy of the top 20 models in the BFCL leaderboard~\footnote{The data is sourced from the BFCL leaderboard update on 2024-09-20, referenced from \url{https://github.com/ShishirPatil/gorilla/blob/e82d4246bec26276cceade9c710df92b9d83420a/data_combined_Sep_20_2024.csv}}. And we compare the GPT-4-turbo and Llama3.1-8B-Instruct on other two benchmarks. The results are shown in Table~\ref{tab:overall} and Table~\ref{tab:overall_other}, where we can have the following observations:

First, ToolACE-DEV, trained using our proposed self-evolution framework, achieves high accuracy at the 8B model scale only, surpassing larger models such as LLaMA-3-70B-Instruct, several closed-source models like Claude, Gemini, and GPT-4o, and models that are specifically fine-tuned for tool invocation. The performance of ToolACE-DEV is on par with that of large MoE models like xLAM-8x22B-r. Besides, ToolACE-DEV shows significant improvements on another two benchmarks compared with Llama-3.1-8B-Instruct. This is attributed to the effectiveness of our training framework.

\begin{figure}[t]
    \centering
    \includegraphics[width=0.96\linewidth]{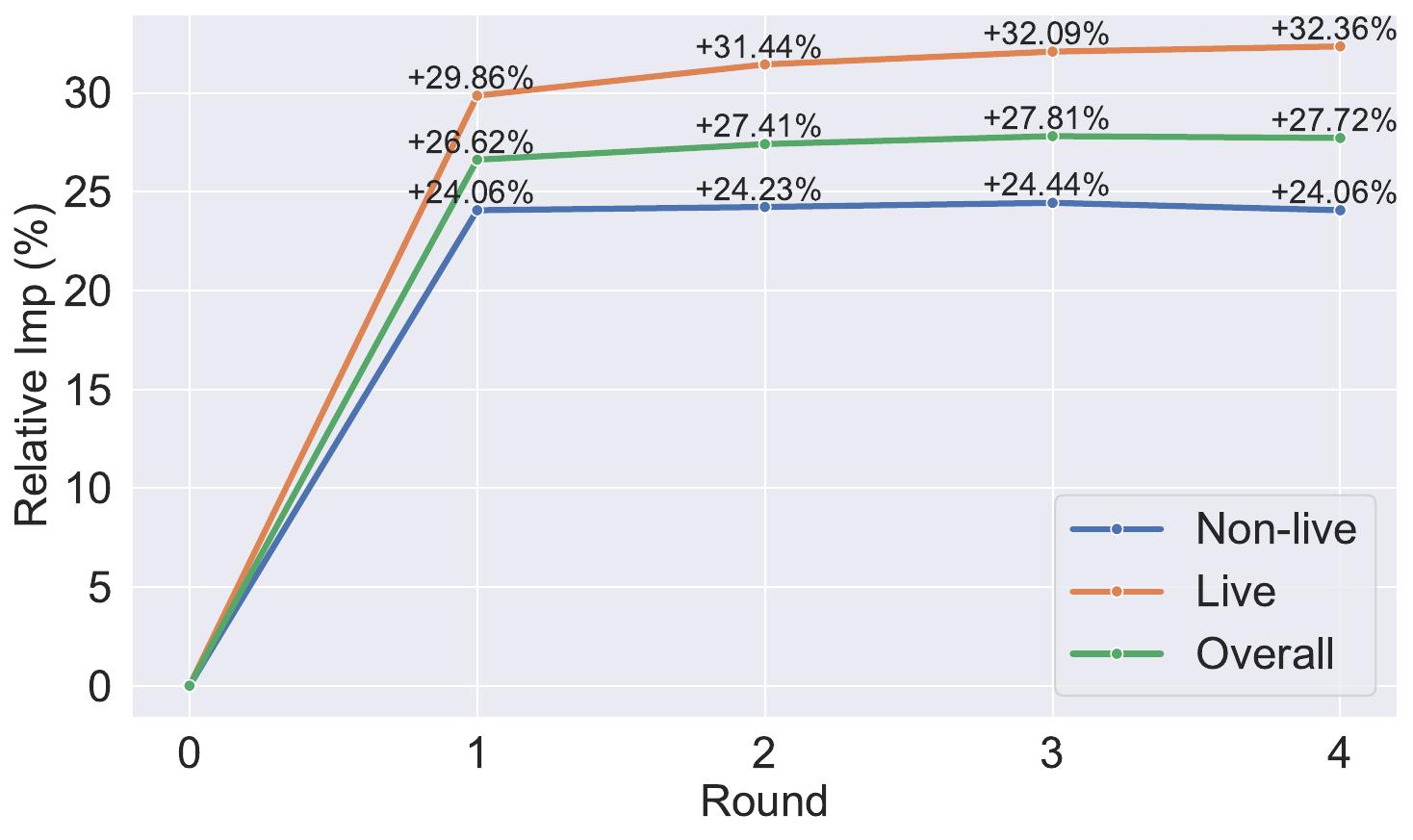}
    \vspace{-0.3cm}
    \caption{Relative Improvements of Self-Evolution. The backbone model is LLaMA-3.1-8B-Instruct.}
    \label{fig:self-evol}
\end{figure}

Second, compared to ToolACE-8B that is finetuned from LLaMA-3.1-8B-Instruct as well, ToolACE-DEV still demonstrates further improvements, achieving higher scores in BFCL. Our new training framework enables ToolACE-DEV to achieve significant improvements with only a minimal amount (20,000) of labeled training data, resulting in reduced training data costs while increasing the data utilization efficiency. Additionally, it fully leverages the model's capability to generate its own data, showcasing that the self-evolution process is as effective as data synthesis with advanced models. This suggests that self-evolution is highly effective for tool-invocation tasks and may become a more efficient approach for data acquisition in the future.

Furthermore, ToolACE-DEV shows a more significant improvement on the more challenging Live subset than on the Non-live subset compared to ToolACE-8B. In the BFCL test set, the candidate tools in the Live category are user-contributed, which increases their authenticity and diversity compared to the Non-live category, making the queries in the Live subset more complex. ToolACE-DEV achieves a larger gain in this more difficult subset, which can be attributed to our tool documentation adaption and query-aware tool generation auxiliary tasks, which improve data utilization. Additionally, the model's self-evolution process generates high-quality training data that is appropriately challenging, rather than simplistic or trivial.

\subsection{Performance of Self-Evolution}\label{sec:exp-evolution}

To evaluate the effectiveness of the model's self-evolution mechanism, we conducted three rounds of self-training using the LLaMA-3.1-8B-Instruct model after it had undergone pre-training phases including Tool Documentation Adaption and Query-Aware Tool Generation and Invocation. In each round of evolution, the model processed 10,000 queries through a two-step generation, producing corresponding candidate tools and tool invocations, as detailed in Section~\ref{sec:self-evol}. The results of each iterative step are presented in Figure~\ref{fig:self-evol}.

It is evident that the scores for Non-live, Live, and Overall metrics consistently improve across iterations, indicating that the model successfully generates informative training data tailored to its current state. Notably, we observe a more pronounced improvement in the more challenging Live scores, suggesting that the data samples generated during self-evolution are appropriately challenging and contain substantial informational value. Additionally, we find that the performance gains from self-evolution diminish as the number of iterations increases. This aligns with conclusions drawn in prior study~\cite{chen2024self}, leading to a hypothesize: the self-iteration process gradually enhances the model's confidence in generating accurate tool invocations, and once the model's confidence becomes sufficiently high, the self-generated data contributes less to further improvements.

\subsection{Ablation Study}

\begin{table}
\centering
\caption{Ablation study on the proposed training objectives. \textbf{Invo}. and \textbf{Gen.} represent the tool invocation and tool generation task, respectively. "w. Adaption" represents the model is trained successively from Adaption.}
\label{tab:ablation}

\setlength{\tabcolsep}{2.5pt}
\begin{tabular}{lccc}
    \toprule
     Model & Non-live & Live & Overall \\ \midrule
     \textbf{Raw} & 72.06 & 56.93 & 64.50 \\
     \textbf{Adaption} & 73.54 & 57.52 & 65.53 \\
     \textbf{Invo.} & 88.96 & 72.73 & 80.85 \\
     \textbf{Invo.} w. Adaption & 89.08 & 73.03 & 81.06 \\
     \textbf{Invo.}+\textbf{Gen.} w. Adaption & \textbf{89.40} & \textbf{73.93 }& \textbf{81.67} \\
     \bottomrule
\end{tabular}
\end{table}

To validate the effectiveness of the training objectives proposed at each stage, we conducted an ablation study to evaluate various variants. Specifically, we compared the following variants:

\textbf{Raw}: The raw model without extra post training.

\textbf{Adaption}: The model undergoes only the first stage of Tool Documentation Adaption training, without any labeled tool invocation data.

\textbf{Invo.}: The model is trained exclusively on the tool invocation portion of the data, using the training objective in Equation~\ref{eq:tool_invocation}.

\textbf{Invo. w. Adaption}: The model first undergoes Tool Documentation Adaption training, followed by training with the tool invocation objective.

\textbf{Invo.+Gen. w. Adaption}: The model first undergoes Tool Documentation Adaption training, then trains both the tool invocation and tool generation objectives simultaneously, optimizing the training objectives in Equation~\ref{eq:tool_invocation} and Equation~\ref{eq:tool_generation}.

The evaluation results for each variant are shown in Table~\ref{tab:ablation}. 
First, the \textbf{Adaption} model shows improvement compared to the Raw model, and \textbf{Invo. w. Adaption} outperforms \textbf{Invo.}, indicating that Tool Documentation Adaption contributes to enhancing the model's understanding of tool definitions and syntax, thereby improving its tool invocation capability. Furthermore, \textbf{Invo.+Gen. w. Adaption} demonstrates a clear advantage among the variants, suggesting that the tool generation task, as an advanced tool-related capability, significantly aids in enhancing the model's tool-related performance.

\begin{figure}[htb]
    \centering
    \subfigure[Various backbone models.]{
        \includegraphics[width=0.9\linewidth]{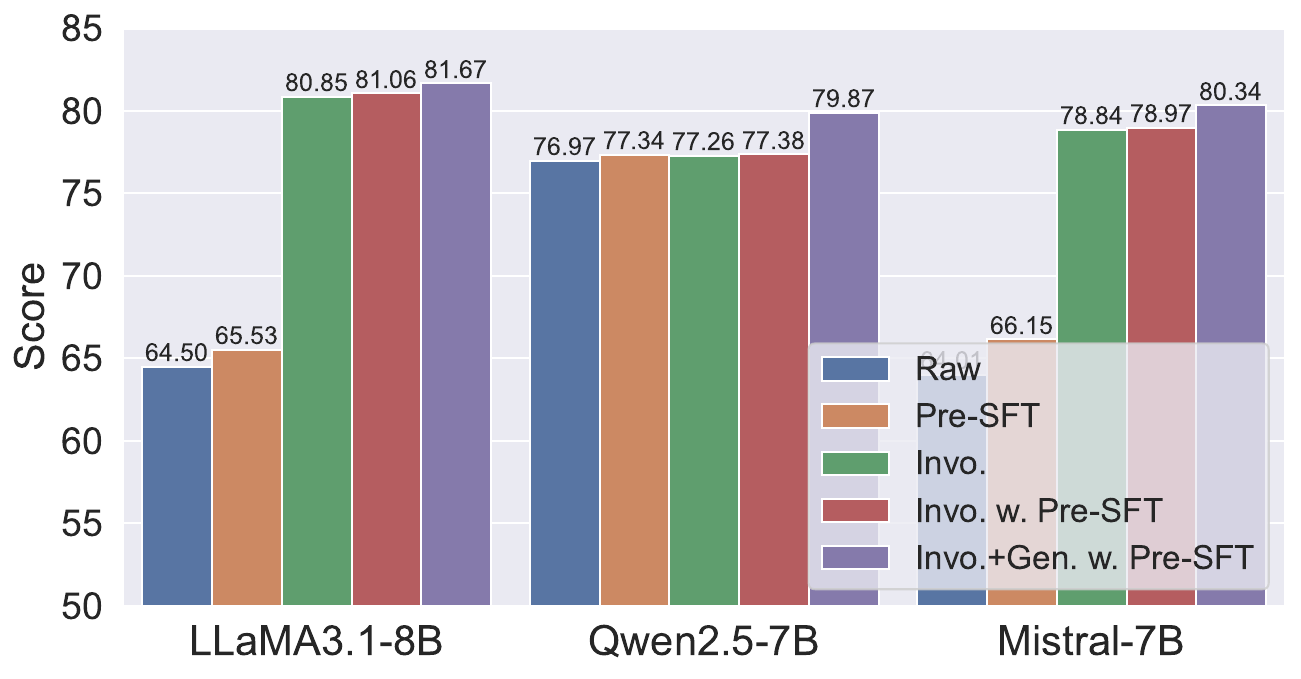}
        \label{fig:ablation-backbone}
    }
    \vspace{-0.3cm}
    \subfigure[Various scales of models.]{
        \includegraphics[width=0.9\linewidth]{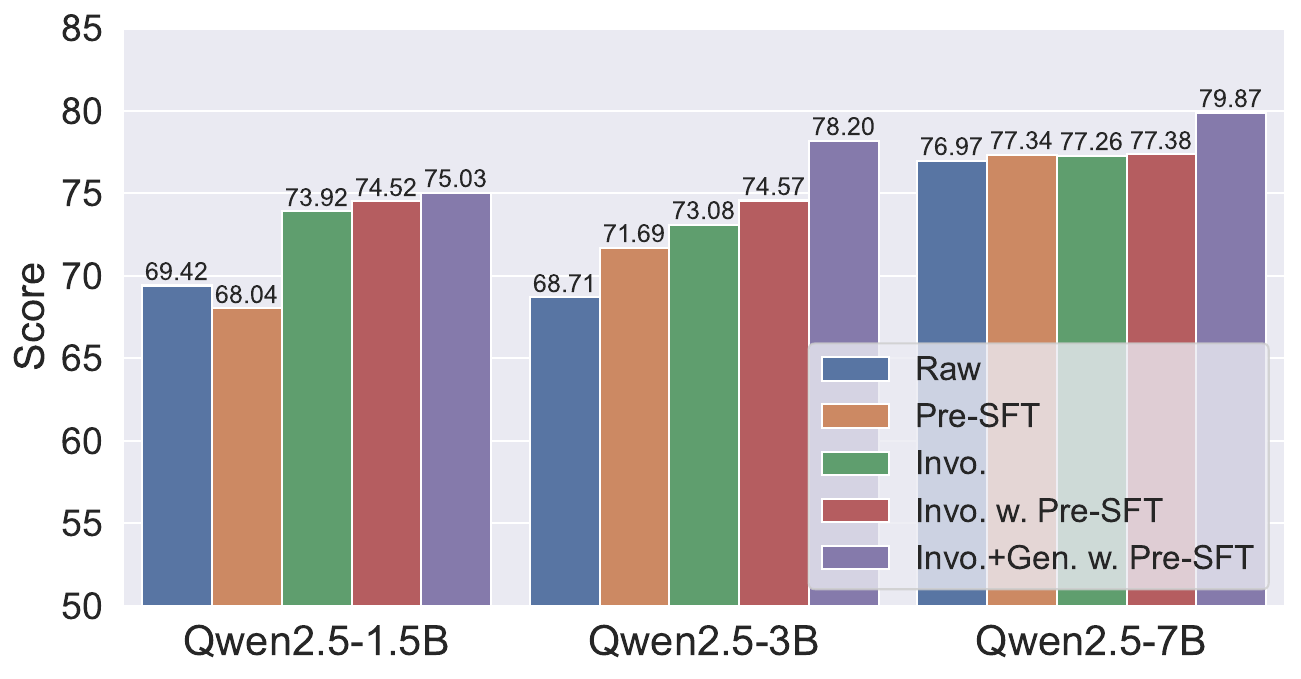}
        \label{fig:ablation-scale}
    }
    \caption{Ablation study of different training objectives on various models.}
    \label{fig:ablation-models}
\end{figure}

\subsection{Effectiveness on various models}
To validate the effectiveness and generality of our proposed framework, we conducted experiments on various models, including models of different parameter scales within the same series and models of similar parameter scales across different series. Specifically, we trained the Qwen2.5-Instruct~\cite{qwen2.5} series models with parameter sizes of 1.5B, 3B, and 7B, as well as the Mistral-v0.3-7B-Instruct~\cite{mistral0.3} and LLaMA-3.1-8B-Instruct models. We evaluated the effectiveness of the proposed training objectives at each stage and the efficacy of the model's self-evolution mechanism.

\noindent \textbf{Effectiveness of training objectives.}
The results of our proposed training objectives across various models are shown in Figure~\ref{fig:ablation-backbone}. As observed, Invo.+Gen. w. Adaption significantly outperforms all other variants, further validating the effectiveness of the proposed Adaption and Generation tasks. Additionally, due to the varying initial tool invocation capabilities of different model backbones, the improvements achieved by Adaption and the invocation tasks differ across models. For instance, Qwen2.5-7B, which exhibits a relatively strong initial tool invocation capability and a better understanding of tools, shows only marginal gains from the Adaption and tool invocation tasks. In contrast, the improvements are more pronounced for LLaMA-3.1-8B and Mistral-v0.3-7B.

\begin{figure}[t]
    \centering
    \subfigure[Various backbone models.\vspace{-0.3cm}]{
        \includegraphics[width=0.9\linewidth]{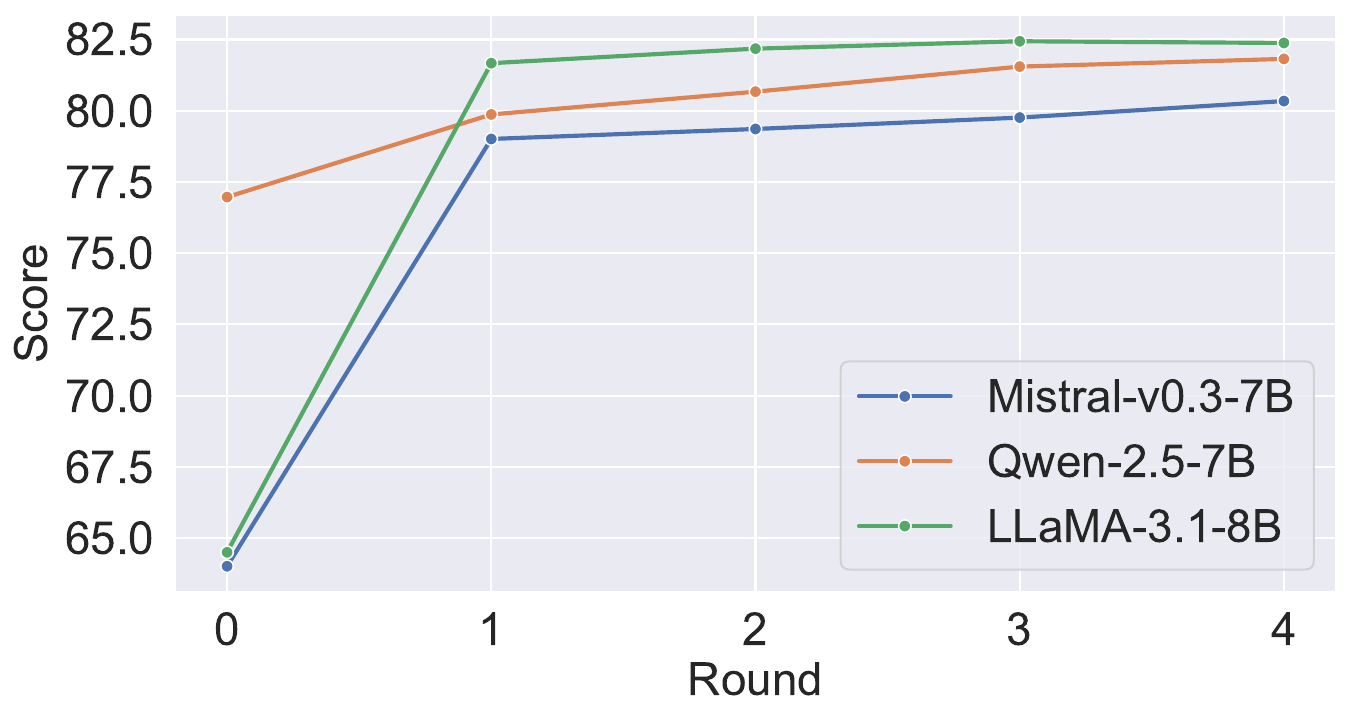}
    }
    \subfigure[Various scales of models.\vspace{-0.3cm}]{
        \includegraphics[width=0.9\linewidth]{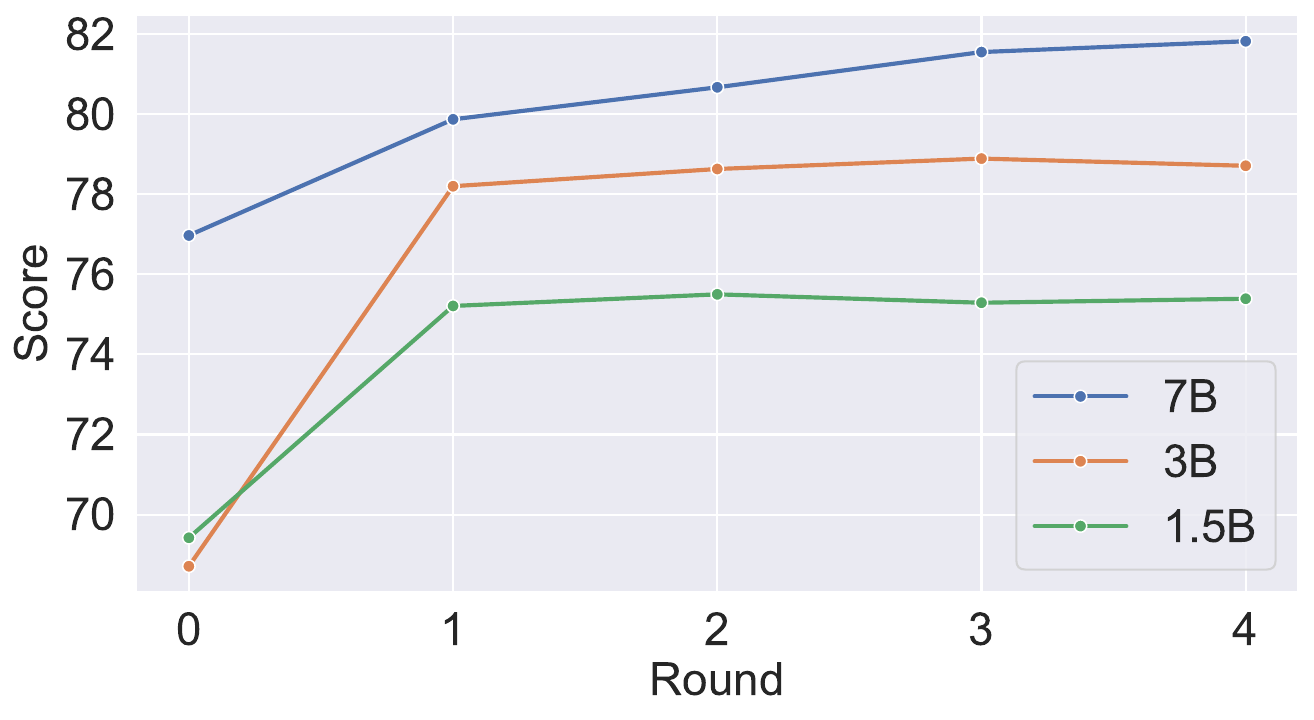}
    }
    \vspace{-0.3cm}
    \caption{Effects of self-evolution on various models.}
    \label{fig:evolution-models}
\end{figure}

As the model size increases, the effects of our proposed training strategy exhibit certain differences, as illustrated in Figure~\ref{fig:ablation-scale}. For smaller models, such as 1.8B model, the weaker instruction-following capabilities result in a tendency to generate tools rather than invoke them when only Tool Documentation Adaption is applied, leading to a decline in tool invocation scores. Additionally, due to the limited number of parameters, the improvements achieved through multitask training in \textbf{Invo.+Gen. w. Adaption} are comparatively smaller. In this case, the tool generation task poses a greater challenge for smaller models, making it more difficult for them to generalize effectively.

\noindent \textbf{Effectiveness of self-evolution.}
The self-evolution performance of different models is illustrated in Figure~\ref{fig:evolution-models}. Our findings are as follows: First, for models with 7-8B parameters, the self-evolution results are consistently positive, aligning with the conclusions drawn in Section~\ref{sec:exp-evolution}. Second, larger models exhibit greater potential for self-evolution. For instance, the 7B models show improvement across all evolution iterations, whereas the 3B models display a slight downward trend in the final iteration, and the 1.5B models exhibit a convergence trend with fluctuations in the last two iterations. This behavior may be attributed to smaller models being more prone to overfitting the training data distribution after 1-2 iterations, which reduces the diversity of the subsequently self-generated data and limits further improvements.

\section{Conclusion}
In this work, we proposed a training framework designed to enhance the tool invocation capabilities of large language models (LLMs), enabling effective self-evolution in tool-related tasks. The training algorithm begins with a Tool Documentation Adaption task to strengthen the model's understanding of tools. Subsequently, we decompose tool-learning data into query-aware tool generation and invocation sub-tasks, empowering the model with the ability to generate tools tailored to specific queries. Building on this foundation, the model can iteratively improve itself by generating data based on given queries. Experimental results demonstrate that our training approach endows the model with self-evolution capabilities, achieves superior tool invocation accuracy compared to all other models of similar scale, and validates the generality of the proposed method across various models.
\section*{Limitations}

First, our experiments were conducted on models up to 7B due to resource constraints, leaving the self-evolution performance of larger models (e.g., 14B, 32B) unexplored. Given existing results, larger models are likely to generate higher-quality data, potentially enhancing self-evolution. Additionally, this work focuses on tool invocation accuracy—selecting the correct tool and providing precise parameters—but does not address retrieving tools from a large-scale tool pool, an important avenue for future research.

\bibliography{mybib}

\appendix
\section{Experimental Details}\label{sec:appd_exp}
\subsection{Benchmark Details}
\textbf{Berkeley Function-Calling Leaderboard (BFCL).} The BFCL~\cite{berkeley-function-calling-leaderboard} benchmark consists of Non-Live and Live categories, where each category comprises single, multiple, parallel and parallel multiple samples. 
\begin{itemize}[leftmargin=*]
    \item Single: A single function evaluation represents the most straightforward yet commonly encountered format, where the user supplies a single JSON function document, and exactly one function call is invoked.
    \item Multiple: The multiple function category involves a user query that triggers a single function call selected from among 2 to 4 available JSON function documents. The model must be capable of determining the most appropriate function to invoke based on the context provided by the user.
    \item Parallel: A parallel function entails the simultaneous invocation of multiple function calls in response to a single user query. The model must determine the number of required function calls, with the user's query potentially consisting of a single sentence or multiple sentences.
    \item Parallel Multiple: Parallel multiple functions combine the concepts of parallel function and multiple function. In this scenario, the model is provided with several function documents, and each corresponding function call may be invoked zero or more times.
\end{itemize}

\noindent \textbf{API-Bank.} API-Bank~\cite{li2023api} is a benchmark designed to evaluate and enhance the tool-augmented capabilities of LLMs, too. It features a runnable evaluation system with 73 API tools and an annotated dataset of 314 dialogues containing 753 API calls, used to assess LLMs' ability to plan, retrieve, and call APIs. In this work, we mainly focus on the tool-invocation task, averaging the correctness of \textit{Call} and \textit{Retrieve+Call} in API-Bank as the overall score.

\noindent \textbf{T-Eval.} T-Eval~\cite{chen-etal-2024-eval} takes several abilities helpful for tool invocation into evaluation, including the  instruction following, planning, reasoning, retrieval, understanding, and review. In this work, we average scores of three tool-invocation task as the overall score: planning, retrieval and understanding, where planning and retrieval represent tool selection and understanding represents parameters filling.

\subsection{Baselines}
We have selected top 20 LLMs on BFCL as baseline models as they show advantaged tool-calling performance, including closed models and open-sourced models. Closed models include GPT-series from OpenAI, Gemini-series from Google and Claude-series from Anthropic. Open-sourced models include general LLMs, such as LLaMA-3-series and mistral-series, and tool-augumented LLMs, such as xLAM-series~\cite{zhang2024xlamfamilylargeaction} and OpenFunctions-series~\cite{patil2023gorilla}. 
For API-Bank and T-Eval, we have compared the state-of-the-art GPT-4-turbo and the LLaMA-3.1-Instruct-8B.

\subsection{Implementation Details}
In the self-evolution stage, we leverage the target model to generate candidate tools and tool invocations by itself. vLLM framework~\cite{kwon2023efficient} is used to accelerate the generation process.  For the generation of the candidate tools, we set the temperature of generation as 1.0, aiming to enhance the diversity of generated tools. For the generation of tool invocation, we utilize the self-consistency decoding strategy, generating 5 solutions for each sample and selecting the solution with the most votes as the final solution.


\section{Prompts}\label{sec:appd_prompt}
In this section, we illustrate all prompts used in the training framework, including the tool invocation task(Figure~\ref{fig:prompt_invocation}), tool documentation pre-SFT task(Figure~\ref{fig:prompt_presft}) and query-aware tool generation task(Figure~\ref{fig:prompt_generation}). In each task, an example comprising of the input prompt and the expected output is detailed.

\begin{figure*}[htb]
    \centering
    \small
\begin{mdframed}
\textsc{\textbf{[system]}}\\
You are a helpful assistant. You can call various user-defined tools to solve users' problems.\\

\# Tools
You may call one or more functions to assist with the user query.\\

You are provided with function signatures within <tools></tools> XML tags:\\
<tools>
\begin{verbatim} 
{
  "name": "Division API", 
  "description": "Divide two time series and return the result.",
  "arguments": {
    "type": "dict",
    "properties": {
      "interval": {
        "description": "Interval between two consecutive points in time series. 
          Supports: `1min`, `5min`, `15min`, `30min`, `45min`, `1h`, `2h`, `4h`, `1day`, `1week`",
        "type": "string"
      }, 
      "symbol": {
        "description": "Instrument symbol, can be any equity, index, ETF, forex or cryptocurrency. 
          E.g. `AAPL`, `EUR/USD`, `ETH/BTC`, ...", 
        "type": "string"
      },
      ...
    "required": ["interval", "symbol"]
  }
  ...
} 
\end{verbatim}
</tools>\\

For each function call, return a json object with function name and arguments within <tool\_call></tool\_call> XML tags:\\
<tool\_call>\\
{"name": <function-name>, "arguments": <args-json-object>}\\
</tool\_call>\\

\textsc{\textbf{[user]}}\\
Could you provide the ratio of the high to the close prices over different intervals for the AAPL symbol? I'd need 1min, 15min and 1h intervals. Make sure to use JSON format for the results.\\

\textsc{\textbf{[Assistant]}}
\color{blue}
\begin{verbatim}
<tool_call>
{
 "name": "Division API", 
 "arguments": {
   "interval": "1min", "symbol": "AAPL", "series_type_1": "high", 
   "series_type_2": "close", "format": "json"
  }
}
</tool_call>
<tool_call>
{
  "name": "Division API", 
  "arguments": {
    "interval": "15min", "symbol": "AAPL", "series_type_1": "high", 
    "series_type_2": "close", "format": "json"
  }
}
</tool_call>
<tool_call>
{
  "name": "Division API", 
  "arguments": {
    "interval": "1h", "symbol": "AAPL", "series_type_1": "high", 
    "series_type_2": "close", "format": "json"
  }
}
</tool_call>
\end{verbatim}
\end{mdframed}
    \caption{Illustration of the tool invocation task. The blue part represents the expected output.}
    \label{fig:prompt_invocation}
\end{figure*}

\begin{figure*}[htb]
    \centering
    \small
\begin{mdframed}
\textsc{\textbf{[system]}}\\
You are a helpful assistant. You are a helpful assistant. You are skilled in generating tool signatures. You can generate high-quality tool signature according to the instruction.\\

Output the tool signature within <tools></tools> XML tags:
\begin{verbatim}
<tools>
{
  "name": "xxx", 
  "description": "xxx",
  "arguments": {
    "type": "dict",
    "properties": {
      "arg1": {
        "type": "xxx",
        "description": "xxx"
      }, 
      "arg2": {...}
    }, 
    "required": ["arg1", ...]
  }, 
  "results": {
    "type": "dict", 
    "properties": {
      "result_1": {"type": "xxx", "description": "xxx"}, "result_2": {...}}
  }
}
</tools>
\end{verbatim}

\textsc{\textbf{[user]}}\\
Generate a tool definition according to the description: Divide two time series and return the result.\\

\textsc{\textbf{[Assistant]}}
\color{blue}
\begin{verbatim}
<tools>
{
  "name": "Division API", 
  "description": "Divide two time series and return the result.",
  "arguments": {
    "type": "dict",
    "properties": {
      "interval": {
        "description": "Interval between two consecutive points in time series. 
          Supports: `1min`, `5min`, `15min`, `30min`, `45min`, `1h`, `2h`, `4h`, `1day`, `1week`",
        "type": "string"
      }, 
      "symbol": {
        "description": "Instrument symbol, can be any equity, index, ETF, forex or cryptocurrency. 
          E.g. `AAPL`, `EUR/USD`, `ETH/BTC`, ...", 
        "type": "string"
      },
      ...
    "required": ["interval", "symbol"]
  }
  ...
}
</tools>
\end{verbatim}
\end{mdframed}
    \caption{Illustration of the tool documentation pre-SFT task. The blue part represents the expected output.}
    \label{fig:prompt_presft}
\end{figure*}

\begin{figure*}[htb]
    \centering
    \small
\begin{mdframed}
\textsc{\textbf{[system]}}\\
You are a helpful assistant. You are a helpful assistant. You are skilled in generating tool signatures. You can generate high-quality tool signature according to the instruction.\\

Output the tool signature within <tools></tools> XML tags:
\begin{verbatim}
<tools>
{
  "name": "xxx", 
  "description": "xxx",
  "arguments": {
    "type": "dict",
    "properties": {
      "arg1": {
        "type": "xxx",
        "description": "xxx"
      }, 
      "arg2": {...}
    }, 
    "required": ["arg1", ...]
  }, 
  "results": {
    "type": "dict", 
    "properties": {
      "result_1": {"type": "xxx", "description": "xxx"}, "result_2": {...}}
  }
}
</tools>
\end{verbatim}

\textsc{\textbf{[user]}}\\
Generate tools to solve the query: Could you provide the ratio of the high to the close prices over different intervals for the AAPL symbol? I'd need 1min, 15min and 1h intervals. Make sure to use JSON format for the results.\\

\textsc{\textbf{[Assistant]}}
\color{blue}
\begin{verbatim}
<tools>
{
  "name": "Division API", 
  "description": "Divide two time series and return the result.",
  "arguments": {
    "type": "dict",
    "properties": {
      "interval": {
        "description": "Interval between two consecutive points in time series. 
          Supports: `1min`, `5min`, `15min`, `30min`, `45min`, `1h`, `2h`, `4h`, `1day`, `1week`",
        "type": "string"
      }, 
      "symbol": {
        "description": "Instrument symbol, can be any equity, index, ETF, forex or cryptocurrency. 
          E.g. `AAPL`, `EUR/USD`, `ETH/BTC`, ...", 
        "type": "string"
      },
      ...
    "required": ["interval", "symbol"]
  }
  ...
}
</tools>
\end{verbatim}
\end{mdframed}
    \caption{Illustration of the query-aware tool generation task. The blue part represents the expected output.}
    \label{fig:prompt_generation}
\end{figure*}


\end{document}